\documentclass[dvipsnames,format=sigconf,anonymous=false,review=false,nonacm]{acmart}

\usepackage{graphicx}
\usepackage{ulem}
\usepackage[utf8]{inputenc}
\usepackage{comment}
\usepackage{array}
\usepackage{longtable}
\usepackage{xspace} 
\usepackage{color}
\usepackage{float}
\usepackage{cancel}
\usepackage{multirow}
\usepackage{amsmath}
\usepackage{mathtools}
\usepackage{soul}
\usepackage{graphicx}
\usepackage{tabularx}
\usepackage{verbatim}
\usepackage{listings}
\usepackage{algorithm}
\usepackage{algorithmicx}
\usepackage{algpseudocode}
\usepackage{multicol}
\usepackage{caption}
\usepackage{titlesec}

\titlespacing{\section}{0pt}{*1.5}{*0.8}
\titlespacing{\subsection}{0pt}{*1.2}{*0.6}

\AtBeginDocument{%
  }

\begin{document}

\title{Surrogate-Assisted Evolution for Efficient Multi-branch Connection Design in Deep Neural Networks}

\author{Fergal Stapleton}
\authornote{Joint first authors}
\orcid{0000-0002-5347-1573}
\affiliation{%
  \institution{Dept. of CS, Hamilton Institute, Maynooth University}
  \city{Kildare}
  \country{Ireland}
}
\email{fergal.stapleton.2020@mumail.ie}

\author{Daniel García Núñez}
\orcid{https://orcid.org/0009-0007-4691-5782}
\affiliation{%
  \institution{Dept. of CS, Hamilton Institute, Maynooth University}
  \city{Kildare}
  \country{Ireland}
}
\email{daniel.garcianunez.2024@mumail.ie}

\author{Yanan Sun}
\orcid{https://orcid.org/0000-0001-6374-1429}
\affiliation{%
\institution{College of Computer Science, Sichuan University}
\city{Chengdu}
\country{China}
}
\email{ysun@scu.edu.cn}

\author{Edgar Galv\'an}
\authornotemark[1]
\affiliation{%
  \institution{Dept. of CS, Hamilton Institute, Maynooth University}
  \city{Kildare}
  \country{Ireland}
}
\email{  edgar.galvan@mu.ie}

\begin{abstract}


State-of-the-art Deep Neural Networks (DNNs) often incorporate multi-branch connections, enabling multi-scale feature extraction and enhancing the capture of diverse features. This design improves network capacity and generalisation to unseen data. However, training such DNNs can be computationally expensive. The challenge is further exacerbated by the complexity of identifying optimal network architectures. To address this, we leverage Evolutionary Algorithms (EAs) to automatically discover high-performing architectures, a process commonly known as neuroevolution. We introduce a novel approach based on Linear Genetic Programming (LGP) to encode multi-branch (MB) connections within DNNs, referred to as NeuroLGP-MB. To efficiently design the DNNs, we use surrogate-assisted EAs. While their application in simple artificial neural networks has been influential, we scale their use from dozens or hundreds of sample points to thousands, aligning with the demands of complex DNNs by incorporating a semantic-based approach in our surrogate-assisted EA. Furthermore, we introduce a more advanced surrogate model that outperforms baseline, computationally expensive, and simpler surrogate models.



\end{abstract}




\maketitle

\section{Introduction}

One of the primary challenges to performing neuroevolution effectively is the significantly high computational overhead required in finding high-quality solutions. It has been noted, that for large DNN models, the cost of training a single epoch can be prohibitive, where this issue is further compounded when many networks are required to train. One way to tackle this expense is through the use of surrogate models, where a distance metric can be used to input the performance of a network without the need to fully train it. One of the major challenges of this approach is comparing networks with different topologies. While distance metrics can be created to compare such topologies using genotypic information, i.e., information pertaining to the structure of the architecture itself, these methods require clever encoding strategies for variable-length architectures. Moving to graph-based topologies further complicates the issue as the genotypic representations would require graph-edit metrics to compare distances~\cite{sanfeliu1983distance}, which can be computationally inefficient to calculate for larger graphs~\cite{ gaier2018data}.    

To this end, we propose Neuro-Linear Genetic Programming Multi-branch (NeuroLGP-MB) which uses Linear Genetic Programming (LGP)~\cite{brameier2007linear} to cleverly encode the diverging and converging paths of a multi-branch topology.  Furthermore, this approach makes use of a semantic-inspired approach to creating distance metrics for surrogate modelling. Semantics, which refer to the output of a program when executed on a given set of inputs, have been widely studied and adopted within the genetic programming (GP) community~\cite{DBLP:conf/gecco/GalvanS19,galvan2022semantics}. However, their study and application in neuroevolution remain underexplored, with only a few exceptions~\cite{Stap2406:NeuroLGP,stapleton2024ola}. The most important aspect of using semantics within this context is that the fixed lengths of these semantic vectors are invariant to architectures of varying length and of different topological complexity, allowing for imputation for a vast array of different topologies, such as those found in multi-branch architectures. 


Thus, the main goal of this work is to find well-performing DNNs by evolving their architectures, which involve multi-branch connections, while doing so efficiently by employing surrogate-assisted EAs. The main contributions of this work are:
i) the use of an ingenious representation based on LGP that allows us to evolve multi-branch connections in DNNs, referred to as NeuroLGP-MB, ii) the use of a semantic-based approach that allows us to make a leap from using dozens/hundreds to thousands of sample points for our Surrogate-assisted EAs model, aligning with current and complex DNNs, iii) the use of a more informative Surrogate-assisted EA model that samples high-quality individuals, which are used to initialise our surrogate model. This method significantly enhances the evolutionary search, while simultaneously reducing the search process significantly, all while still attaining high performance compared to the other three approaches used in this work: the simple Surrogate-assisted EA model, baseline, and expensive approaches, iv) and an analysis on the elite members of the population, looking at how depth and complexity changed over time, demonstrating the topological preferences for the NeuroLGP-MB method on the tested datasets.

\section{Methodology}

\subsection{Kriging Partial Least Squares}
\label{chp:background:sub:kriging}




In the context of neuroevolution, we can define a solution sample $\mathrm{x}$ as having the semantic or phenotypic behaviour of the $i^{th}$ program (in this case network) such that
$\mathrm{x}_i = s(p_i)$, where the semantics $s(p)$ of a program $p$ is the vector of values from the output. We define our distance $\mathcal{D}$ as a phenotypic distance, as seen in Eq.~\ref{eqn::kernel_dist3} 

\begin{equation}
\mathcal{D}(\mathrm{x}_i, \mathrm{x}_j) = \mathcal{D}(s(p_i), s(p_j)))
\label{eqn::kernel_dist3}
\end{equation}

\noindent where the distance metric is dependent on the outputs of each network. In this work, our semantic vector is a flattened vector containing the output of the nodes at the final softmax layer for all data instances. As such, our semantic vector length is given as the number of images of the test dataset $\times$ the number of classes.

Kriging is an interpolation-based technique that assumes spatial correlation exists between known data points, based on the distance, and variation between these points. In this work, we use a variation of Kriging, known as Kriging Partial Least squares (KPLS) to tackle the prohibitive cost of estimating $\theta$ parameters, which typically requires maximum likelihood estimation. In essence, we use the KPLS approach to estimate the fitness of unknown networks using the semantic vectors as detailed above. For full details of the KPLS method, refer to~\cite{bouhlel2016improving}.

\subsection{NeuroLGP-Multi-Branch}

\begin{figure}[tbp]

    \centering
    \begin{minipage}{0.24\textwidth}
        \centering
        \resizebox{\textwidth}{!}{
        \begin{tabular}{|c|l|l|l|l|}
        \hline
         & \textbf{Operand} & \multicolumn{3}{c|}{\textbf{Register}} \\ \hline
        \hline
        \textbf{Id} & \textbf{Layer} & \textbf{O} & \textbf{I$_2$} & \textbf{I$_2$} \\ \hline
        17 & CONV\_64\_3x3 & r1 & r0 & \\ \hline
        \textbf{16} & \textbf{CONV\_128\_5x5} & \textcolor{cyan}{\textbf{r5}} & \textcolor{cyan}{\textbf{r6*}}& \\ \hline
        15 & DROPOUT\_0.3 & r4 & r1 & \\ \hline
        14 & AVGPOOL\_3x3 & r1 & r1 & \\ \hline
        13 & CONV\_64\_5x5 & r4 & r2 & \\ \hline
        12 & CONV\_128\_3x3 & r4 & r1 & \\ \hline
        11 & CONV\_64\_3x3 & r1 & r5 & \\ \hline
        10 & BATCH\_NORM & r1 & r5 & \\ \hline
        \textbf{9} & \textbf{DROPOUT\_0.5} & \textcolor{cyan}{\textbf{r1}} & \textcolor{cyan}{\textbf{r5}} & \\ \hline
        \textbf{8} & \textbf{MAXPOOL\_3x3} & \textcolor{magenta}{\textbf{r3}} & \textcolor{magenta}{\textbf{r2*}} & \\ \hline
        7 & CONV\_64\_3x3 & r5 & r5 & \\ \hline
        \textbf{6} & \textbf{CONCAT} & \textcolor{blue}{\textbf{r2}} & \textcolor{cyan}{\textbf{r1}} & \textcolor{magenta}{\textbf{r3}} \\ \hline
        5 & CONV\_64\_5x5 & r0 & r5 & \\ \hline
        4 & AVGPOOL\_5x5 & r4 & r2 & \\ \hline
        3 & CONV\_128\_5x5 & r3 & r0 & \\ \hline
        2 & CONV\_32\_3x3 & r1 & r1 & \\ \hline
        \textbf{1} & \textbf{BATCH\_NORM} & \textcolor{blue} {\textbf{r0}} & \textcolor{blue}{\textbf{r2}} & \\ \hline \end{tabular}}
    \end{minipage}%
    \hfill
    \begin{minipage}{0.225\textwidth}
        \centering
        \includegraphics[width=\textwidth]{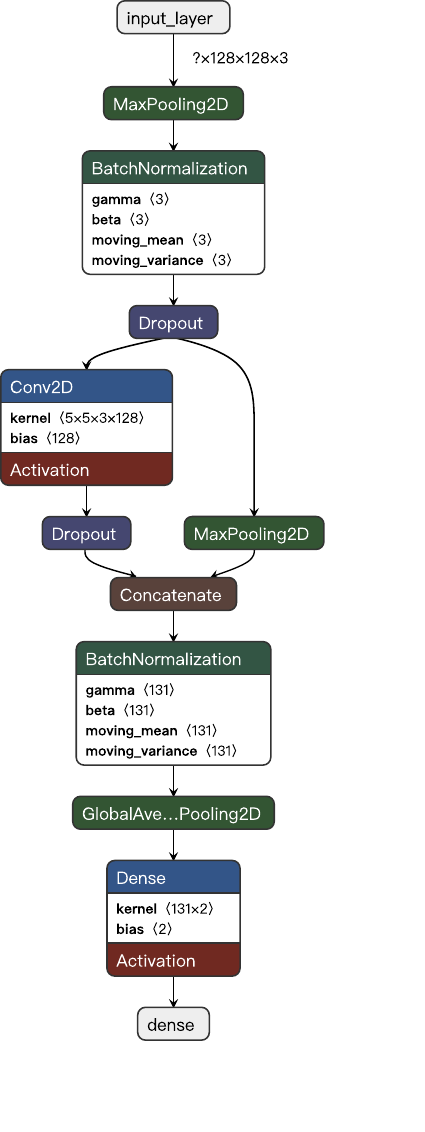}
    \vspace{-35pt}\end{minipage}      
  \caption{Left: Genotypic representation of the DNN architecture. Bold denotes effective code. Registers are colour-coded (see text for explanation). Right: Graphical representation of the same genotype, showing how the architecture is constructed from the effective code.} \label{fig:combined_geno_pheno}
\end{figure}

 On the left of Figure~\ref{fig:combined_geno_pheno}, we demonstrate the genotypic representation for the NeuroLGP-MB approach. The genotype contains effective (bold) and non-effective code (non-bold). While the ordering of the layers is interpreted from a top-down perspective, i.e., layers at the top will appear first after the input, the network graph is constructed in reverse (hence, the Id column is in descending order). As such when constructing the architecture, we start on the Id 1 for BATCH\_NORM which uses the special output operand r[0] to denote the starting point. The motivation is that the CONCAT layer (Id 6) can be used to denote when a split in the graph occurs and as such we traverse the graph in reverse order (denoted by light blue and magenta registers). The right of Fig.~\ref{fig:combined_geno_pheno} shows the phenotypic representation of the genotype defined in the table. The registers highlighted in light blue in this table correspond to the left branch of this figure and registers highlighted in magenta correspond to the right branch. Additional non-evolved operations are required to ensure the layers are similar sizes for the concatenation layer and occur on the 3rd channel. GlobalAveragePooling2D and Dense layer are also not expressed in genotype, as such our evolutionary process is primarily focussed on feature extraction.

\subsection{Surrogate Model Management}
\label{sec::surrmodmang}

\begin{figure}[tb]
 
  \centering
\includegraphics[width=0.49\textwidth]{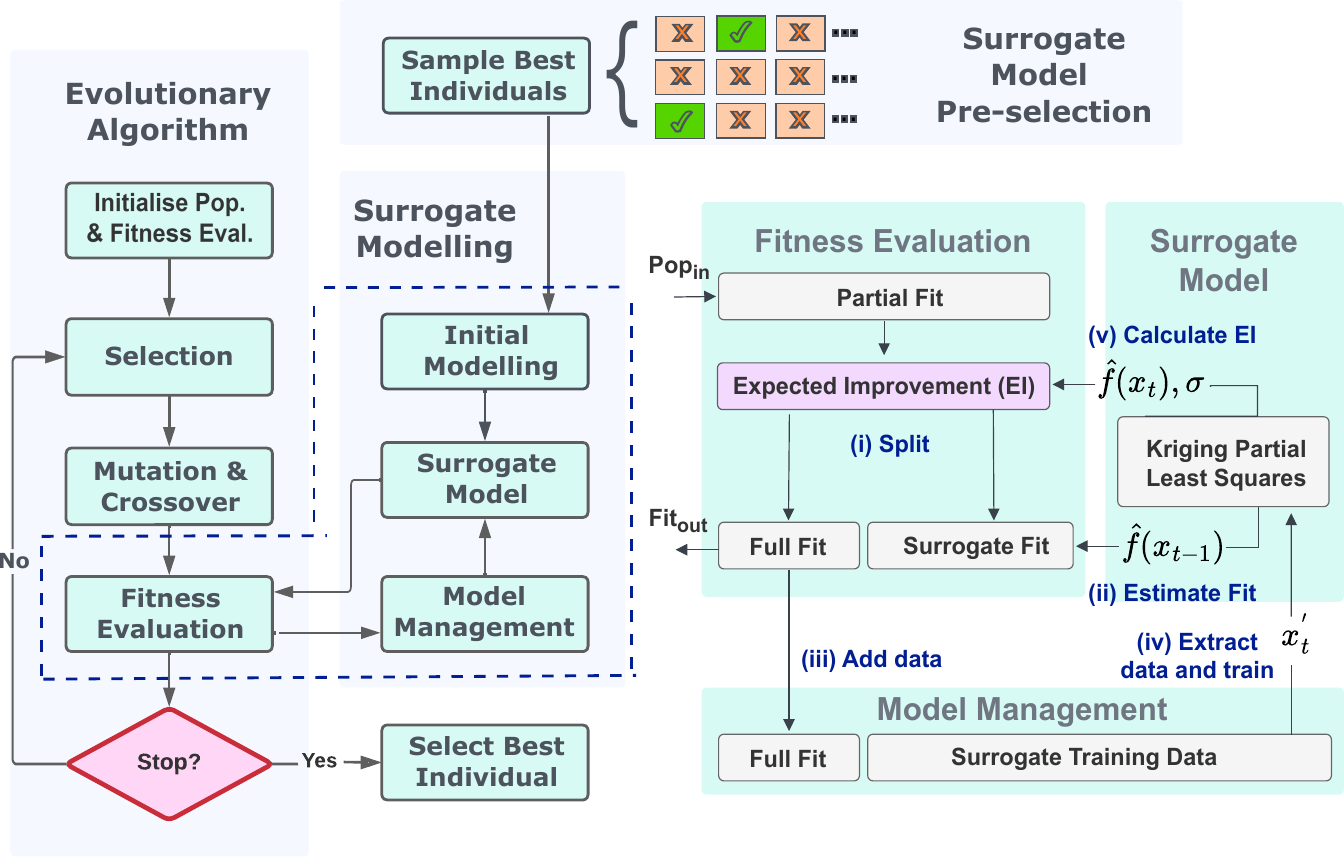}
  \vspace{-15pt}
  \caption{\textit{Left:} Diagram showing the interplay between a typical evolutionary algorithm and a surrogate model approach. \textit{Right:} The surrogate model management strategy is shown on a more granular level. \textit{Top:} Pre-selection approach used to initialise the surrogate model.}

  \label{fig:surrogate_overview}
\end{figure}

Fig.~\ref{fig:surrogate_overview} summarises the surrogate model management strategy. The left hand side of the figure shows the interplay between the evolutionary algorithm and the surrogate model, however, we have highlighted the steps controlled by the model management strategy (dark blue dashed-line). The right side of the figure looks at the model management strategy at a more granular level. On the top of the figure we demonstrate the pre-selection approach, which samples individuals based on their performance and is used to inform the initial surrogate model. In the original surrogate approach the 1st generation is instead informed with randomly selected individuals. The diagram has been annotated with five key steps (dark blue text). The annotations are: (i) \textit{Split:} first, the population is split with a 40/60 split for individuals to be fully evaluated \textit{vs.} the partially trained individuals, respectively, (ii) \textit{Estimate fitness:} the fitness is estimated using the KPLS approach as informed by the previous generation, (iii) \textit{Add data:} the semantic distance vector for the fully evaluated portion of the population is added to the surrogate training data, (iv) \textit{Extract data and train:} the phenotypic distance vectors are collectively used to train the KPLS approach as detailed in Section~\ref{chp:background:sub:kriging}, and (v) \textit{Calculate EI:} the EI criteria is calculated for the incoming population of individuals.

\section{Experimental details}
\label{sec::exp}

\subsection{Datasets}

We used three datasets, two variants from the Breast Cancer \sloppy{Histopathological} image classification dataset (BreakHis)~\cite{spanhol2016breast} and the Chest X-Ray binary classification dataset for patients with pneumonia~\cite{kermany2018labeled}. More details on the datasets can be found in their respective papers~\cite{spanhol2016breast}~\cite{kermany2018labeled}, but the number of images ranges from 1.5 to 5 thousand images. Training, validation, test set were used for network evaluation and a further test set was used for reporting ($\sim$15\% of dataset size). The images have been down-scaled to 128x128 pixels. Data-augmentation techniques and up-sampling techniques are used to ensure our classes are balanced (training only).
We employ 4 experimental methods in this work: i) a \textbf{Baseline} approach which is similar to a random search approach, in that, it structures the network architecture layers in a random order. These networks are not evolved. All networks are fully trained to the full number of epochs (30). ii) the \textbf{Expensive} approach, which uses a novel encoding to evolve the structure of our networks, training to the full number of epochs (Pop. size = 50, Gen. size =15), iii) The \textbf{Surrogate} approach where we integrate the surrogate modelling strategy outlined in Section~\ref{sec::surrmodmang}. The surrogate model is informed using partially trained networks (10 epochs) for 60\% of the population size.
iv) the \textbf{Surrogate-PS} variant as outlined in Section~\ref{sec::surrmodmang}.





\section{Results}


\newcommand{\trainfortymean}{0.822}
\newcommand{\trainfortystd}{0.018}
\newcommand{\valfortymean}{0.862}
\newcommand{\valfortystd}{0.026}
\newcommand{\testfortymean}{0.837}
\newcommand{\testfortystd}{0.022}
\newcommand{\testfortytwomean}{0.887}
\newcommand{\testfortytwostd}{0.012}

\newcommand{\trainsurfortymean}{0.865}
\newcommand{\trainsurfortystd}{0.054}
\newcommand{\valsurfortymean}{0.898}
\newcommand{\valsurfortystd}{0.032}
\newcommand{\testsurfortymean}{0.873}
\newcommand{\testsurfortystd}{0.044}
\newcommand{\testsurfortytwomean}{0.904}
\newcommand{\testsurfortytwostd}{0.029}

\newcommand{\trainsurpsfortymean}{0.907}
\newcommand{\trainsurpsfortystd}{0.061}
\newcommand{\valsurpsfortymean}{0.910}
\newcommand{\valsurpsfortystd}{0.032}
\newcommand{\testsurpsfortymean}{0.889}
\newcommand{\testsurpsfortystd}{0.040}
\newcommand{\testsurpsfortytwomean}{0.919}
\newcommand{\testsurpsfortytwostd}{0.032}

\newcommand{\trainexpfortymean}{0.897}
\newcommand{\trainexpfortystd}{0.032}
\newcommand{\valexpfortymean}{0.894}
\newcommand{\valexpfortystd}{0.037}
\newcommand{\testexpfortymean}{0.873}
\newcommand{\testexpfortystd}{0.038}
\newcommand{\testexpfortytwomean}{0.907}
\newcommand{\testexpfortytwostd}{0.029}

\newcommand{\traintwohundredmean}{0.944}
\newcommand{\traintwohundredstd}{0.017}
\newcommand{\valtwohundredmean}{0.883}
\newcommand{\valtwohundredstd}{0.033}
\newcommand{\testtwohundredmean}{0.890}
\newcommand{\testtwohundredstd}{0.008}
\newcommand{\testtwohundredtwomean}{0.897}
\newcommand{\testtwohundredtwostd}{0.014}

\newcommand{\trainsurtwohundredmean}{0.935}
\newcommand{\trainsurtwohundredstd}{0.044}
\newcommand{\valsurtwohundredmean}{0.890}
\newcommand{\valsurtwohundredstd}{0.035}
\newcommand{\testsurtwohundredmean}{0.905}
\newcommand{\testsurtwohundredstd}{0.011}
\newcommand{\testsurtwohundredtwomean}{0.919}
\newcommand{\testsurtwohundredtwostd}{0.031}

\newcommand{\trainsurpstwohundredmean}{0.964}
\newcommand{\trainsurpstwohundredstd}{0.010}
\newcommand{\valsurpstwohundredmean}{0.923}
\newcommand{\valsurpstwohundredstd}{0.025}
\newcommand{\testsurpstwohundredmean}{0.909}
\newcommand{\testsurpstwohundredstd}{0.020}
\newcommand{\testsurpstwohundredtwomean}{0.939}
\newcommand{\testsurpstwohundredtwostd}{0.010}

\newcommand{\trainexptwohundredmean}{0.941}
\newcommand{\trainexptwohundredstd}{0.021}
\newcommand{\valexptwohundredmean}{0.912}
\newcommand{\valexptwohundredstd}{0.024}
\newcommand{\testexptwohundredmean}{0.911}
\newcommand{\testexptwohundredstd}{0.013}
\newcommand{\testexptwohundredtwomean}{0.930}
\newcommand{\testexptwohundredtwostd}{0.030}

\newcommand{\trainxraymean}{0.796}
\newcommand{\trainxraystd}{0.119}
\newcommand{\valxraymean}{0.894}
\newcommand{\valxraystd}{0.016}
\newcommand{\testxraymean}{0.914}
\newcommand{\testxraystd}{0.020}
\newcommand{\testxraytwomean}{0.912}
\newcommand{\testxraytwostd}{0.015}

\newcommand{\trainsurxraymean}{0.863}
\newcommand{\trainsurxraystd}{0.051}
\newcommand{\valsurxraymean}{0.900}
\newcommand{\valsurxraystd}{0.018}
\newcommand{\testsurxraymean}{0.914}
\newcommand{\testsurxraystd}{0.018}
\newcommand{\testsurxraytwomean}{0.909}
\newcommand{\testsurxraytwostd}{0.021}

\newcommand{\trainsurpsxraymean}{0.865}
\newcommand{\trainsurpsxraystd}{0.067}
\newcommand{\valsurpsxraymean}{0.906}
\newcommand{\valsurpsxraystd}{0.011}
\newcommand{\testsurpsxraymean}{0.921}
\newcommand{\testsurpsxraystd}{0.006}
\newcommand{\testsurpsxraytwomean}{0.914}
\newcommand{\testsurpsxraytwostd}{0.014}

\newcommand{\trainexpxraymean}{0.856}
\newcommand{\trainexpxraystd}{0.107}
\newcommand{\valexpxraymean}{0.895}
\newcommand{\valexpxraystd}{0.023}
\newcommand{\testexpxraymean}{0.920}
\newcommand{\testexpxraystd}{0.030}
\newcommand{\testexpxraytwomean}{0.917}
\newcommand{\testexpxraytwostd}{0.016}

\subsection{Performance and Surrogate Analysis}
\label{chp:third:results:surrogate}

Comparing the test accuracy based on four runs for BreakHis $\times$40 we found that the surrogate-PS (underlined) had the highest average accuracy, \underline{$\testsurpsfortytwomean \pm
\testsurpsfortytwostd$}, \textit{vs.} $\testfortytwomean \pm
\testfortytwostd$, $\testexpfortytwomean \pm
\testexpfortytwostd$ and $\testsurfortytwomean \pm
\testsurfortytwostd$, for baseline, expensive and surrogate models, respectively. Again, for the BreakHis $\times$200, we found that surrogate-PS outperforms the other methods with an accuracy of \underline{$\testsurpstwohundredtwomean \pm
\testsurpstwohundredtwostd$} \textit{vs.} $\testtwohundredtwomean \pm
\testtwohundredtwostd$, $\testexptwohundredtwomean \pm
\testexptwohundredtwostd$ and $\testsurtwohundredtwomean \pm
\testsurtwohundredtwostd$, for baseline, expensive and surrogate models, respectively. In the case of Chest X-Ray the accuracies were all relatively similar, \underline{$\testsurpsxraytwomean \pm
\testsurpsxraytwostd$,} \textit{vs.} $\testxraytwomean \pm
\testxraytwostd$,  $\testexpxraytwomean \pm
\testexpxraytwostd$,  $\testsurxraytwomean \pm
\testsurxraytwostd$ for surrogate-PS, baseline, expensive and surrogate, respectively. Barring the Chext X-Ray dataset, the surrogate-PS method performed the best.

\newcommand{\mseXfortyvtwo}{0.0067}
\newcommand{\mseXfortyvtwops}{0.0035}
\newcommand{\mseXtwohundredvtwo}{0.0017}
\newcommand{\mseXtwohundredvtwops}{0.0031}
\newcommand{\mseXchestxrayvtwo}{0.0372}
\newcommand{\mseXchestxrayvtwops}{0.0173}
\newcommand{\kendallXfortyvtwo}{0.6536}
\newcommand{\kendallXfortyvtwops}{0.6480}
\newcommand{\kendallXtwohundredvtwo}{0.6480}
\newcommand{\kendallXtwohundredvtwops}{0.7435}
\newcommand{\kendallXchestxrayvtwo}{0.6298}
\newcommand{\kendallXchestxrayvtwops}{0.6790}
\newcommand{\codXfortyvtwo}{0.6239}
\newcommand{\codXfortyvtwops}{0.6185}
\newcommand{\codXtwohundredvtwo}{0.9373}
\newcommand{\codXtwohundredvtwops}{0.7458}
\newcommand{\codXchestxrayvtwo}{0.4536}
\newcommand{\codXchestxrayvtwops}{0.6082}


\begin{table}[tb]
  \centering
    \caption{Average MSE, Kendall's Tau and R$^2$ for different datasets. The naming convention has been shortened for ease of read, such that, the original surrogate model is denoted as `Sur' where the pre-selection method is denoted as `Sur-PS'.}
\resizebox{0.45\textwidth}{!}{%
  \begin{tabular}{lcccccc}
    \hline
    \multirow{3}{*}{Metric} & 
    \multicolumn{6}{c}{Datasets} \\
    \cline{2-7}
     & \multicolumn{2}{c}{$\times$40} & \multicolumn{2}{c}{$\times$200}  & \multicolumn{2}{c}{Chest X-Ray} \\
    
    \cline{2-7}
     & Sur & Sur-PS & Sur & Sur-PS  & Sur & Sur-PS \\
    \hline
    MSE & \mseXfortyvtwo & \mseXfortyvtwops & \mseXtwohundredvtwo & \mseXtwohundredvtwops & \mseXchestxrayvtwo &  \mseXchestxrayvtwops \\
    Kendall's Tau &\kendallXfortyvtwo & \kendallXfortyvtwops & \kendallXtwohundredvtwo & \kendallXtwohundredvtwops & \kendallXchestxrayvtwo & \kendallXchestxrayvtwops \\
    R$^2$ & \codXfortyvtwo & \codXfortyvtwops & \codXtwohundredvtwo & \codXtwohundredvtwops &  \codXchestxrayvtwo & \codXchestxrayvtwops \\
    \hline
  \end{tabular}
  }
\label{tab:surrogate_effectiveness_v2}
\end{table}

Table~\ref{tab:surrogate_effectiveness_v2} summarises the effectiveness of both the surrogate (Sur) model and the pre-selection surrogate (Sur-PS) model for each of the datasets for the 3 metrics. If we compare the metrics for the base surrogate and pre-selection variants against each other (sur \textit{vs.} sur-PS), we can see that, on the surface, the metrics tend to vary as to which is preferable, for instance, if we look at $\times$40 we can see that surrogate-PS has a lower MSE of $\mseXfortyvtwops$ \textit{vs.} $\mseXfortyvtwo$ (value of 0 represents ideal) but the Kendell's Tau and R$^2$ are preferable for the surrogate model with $\kendallXfortyvtwo$ \textit{vs.} $\kendallXfortyvtwops$ and $\codXfortyvtwo$ \textit{vs.}
$\codXfortyvtwops$ (value of 1 represents ideal in both cases). Only in the case of Chest X-Ray do the metrics seem to be consistently preferable. In general, we can say the surrogate model performed robustly.


Now, if we focus specifically on the datasets, we can see that the Kendall's Tau and R$^2$ values show a relative moderate to high correlation, with values typically above 0.6 (with the exception of Chest X-Ray surrogate model that has a R$^2$ value of  $\codXchestxrayvtwo$). The MSE values are low (preferable) for BreakHis $\times$40 and BreakHis $\times$200, however for Chest X-Ray, the MSE values are an order of magnitude higher. This implies that even though there is a high correlation between predicted versus actual results for Chest X-ray there is a high degree of variance.

\newcommand{\timexfortybaselinevtwoMean}{26.46}
\newcommand{\timexfortybaselinevtwoStd}{0.00}
\newcommand{\timexfortyexpensivevtwoMean}{25.46}
\newcommand{\timexfortyexpensivevtwoStd}{0.05}
\newcommand{\timexfortysurrogatevtwoMean}{19.55}
\newcommand{\timexfortysurrogatevtwoStd}{0.23}
\newcommand{\timexfortysurrogatepsvtwoMean}{21.45}
\newcommand{\timexfortysurrogatepsvtwoStd}{0.24}
\newcommand{\timexfortysurrogatevtwopercMean}{23.2}
\newcommand{\timexfortysurrogatevtwopercStd}{1.2}
\newcommand{\timexfortysurrogatepsvtwopercMean}{15.8}
\newcommand{\timexfortysurrogatepsvtwopercStd}{1.2}

\newcommand{\timextwohundredbaselinevtwoMean}{27.10}
\newcommand{\timextwohundredbaselinevtwoStd}{0.07}
\newcommand{\timextwohundredexpensivevtwoMean}{27.32}
\newcommand{\timextwohundredexpensivevtwoStd}{0.02}
\newcommand{\timextwohundredsurrogatevtwoMean}{20.42}
\newcommand{\timextwohundredsurrogatevtwoStd}{0.74}
\newcommand{\timextwohundredsurrogatepsvtwoMean}{22.77}
\newcommand{\timextwohundredsurrogatepsvtwoStd}{0.30}
\newcommand{\timextwohundredsurrogatevtwopercMean}{25.3}
\newcommand{\timextwohundredsurrogatevtwopercStd}{3.6}
\newcommand{\timextwohundredsurrogatepsvtwopercMean}{16.6}
\newcommand{\timextwohundredsurrogatepsvtwopercStd}{1.3}

\newcommand{\timexxraybaselinevtwoMean}{21.03}
\newcommand{\timexxraybaselinevtwoStd}{0.04}
\newcommand{\timexxrayexpensivevtwoMean}{21.13}
\newcommand{\timexxrayexpensivevtwoStd}{0.02}
\newcommand{\timexxraysurrogatevtwoMean}{14.86}
\newcommand{\timexxraysurrogatevtwoStd}{0.16}
\newcommand{\timexxraysurrogatepsvtwoMean}{18.63}
\newcommand{\timexxraysurrogatepsvtwoStd}{0.13}
\newcommand{\timexxraysurrogatevtwopercMean}{29.7}
\newcommand{\timexxraysurrogatevtwopercStd}{1.1}
\newcommand{\timexxraysurrogatepsvtwopercMean}{11.8}
\newcommand{\timexxraysurrogatepsvtwopercStd}{0.7}



\begin{figure*}[t]
    \centering
    
    \begin{minipage}{0.31\textwidth}
        \centering
        \includegraphics[width=\linewidth]{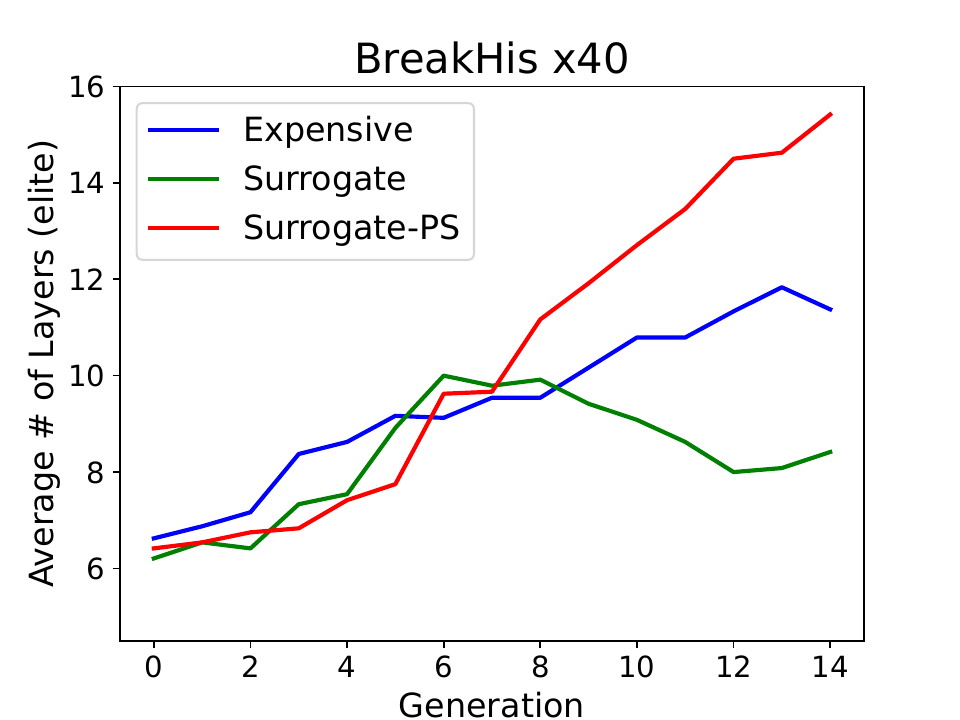
} 
      
        \label{fig:x40depth}
    \end{minipage}
    \hfill 
    \begin{minipage}{0.31\textwidth}
        \centering
        \includegraphics[width=\linewidth]{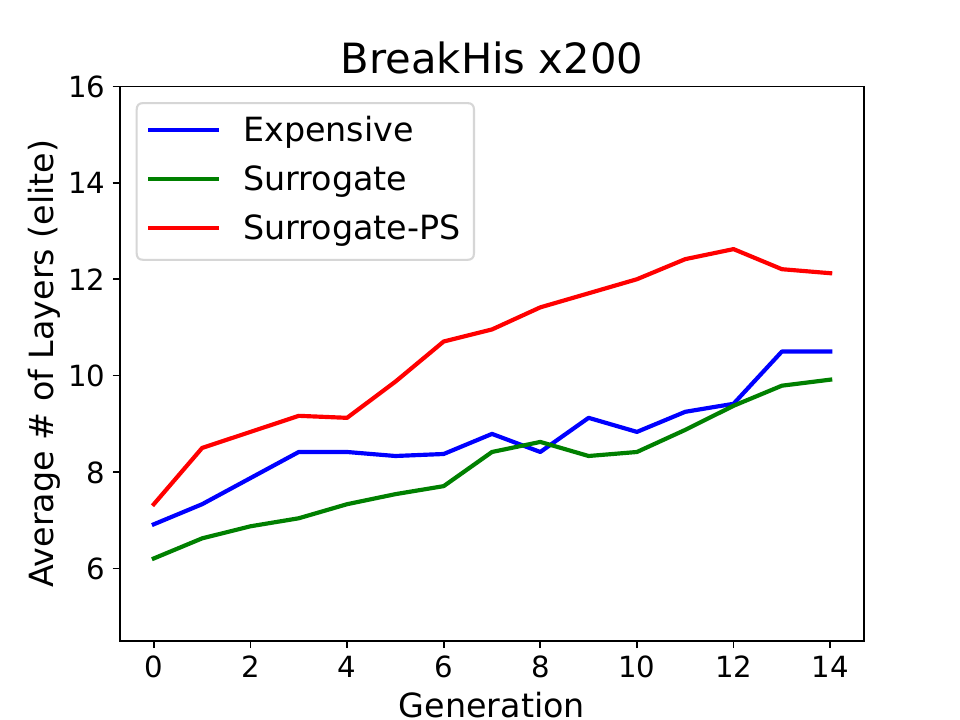
} 

        \label{fig:x200depth}
    \end{minipage}
    \hfill
    \begin{minipage}{0.31\textwidth}
        \centering
    \includegraphics[width=\linewidth]          {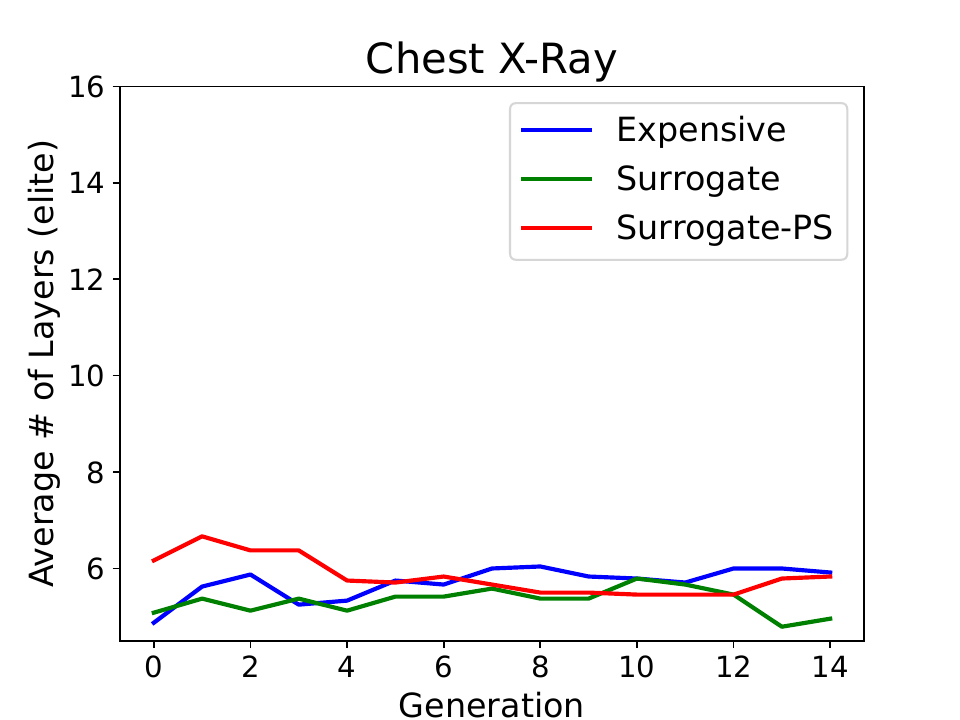
} 

    \end{minipage}

    \caption{Average number of layers for expensive, surrogate and surrogate-PS across 15 generation for each dataset.}
    \label{fig:x40andx200depthandXray}
\end{figure*}

A time analysis showed that there is a relatively high variance in the number of GPU hours of the surrogate and surrogate-PS models, but both approaches are significantly faster than the expensive model. The percentage reduction compared to the expensive model for the original surrogate model was 23.2 - 28.7\% across the three datasets. However, for the surrogate-PS model, considering the significant overhead in evaluating individuals for this method (150 individuals trained to 10 epochs), there was still a significant time saving ranging from 11.8 - 16.6\%. Based on the previous results, where the surrogate-PS model was found to perform the best for BreakHis $\times$40 and BreakHis $\times$200, there is a clear advantage in considering the surrogate-PS model.

\subsection{Network Depth and Complexity}


The depth analysis was performed on the elite population, which was 20\% of the total population size, as such, each elite population represents the 6 best individuals (out of 30) from each generation per run. Fig.~\ref{fig:x40andx200depthandXray} shows the change in average number of layers of the elite population across 15 generations, for the three datasets. Looking at the left and centre figure for BreakHis $\times$40 and BreakHis $\times$200 datasets, we can see that in both datasets, surrogate-PS (red line) tends to generate individuals with the largest number of layers, or greatest depth, followed by the expensive (blue line) and then the surrogate (green line). These results show that as the accuracy in each generation increased, the
depth of the networks found also increased in general. Looking at the right-hand figure, focusing now on the Chest X-Ray dataset, we can see that on average the number of layers is not changing over time for the expensive, surrogate and surrogate-PS models. There may be a number of contributing factors as to why the depth does not increase. There is a strong indication from our previous analysis that random sampling, akin to the baseline model, is capable of finding competitive networks.

As the concatenation layer was the contributing factor in determining network topological complexity, we performed an analysis of their proportions relative to the elite population members in the final generation specifically. We found that typically the number of concatenation layers ranged from 1 in 33 layers to 1 in 8 layers, depending on the dataset and run. This showed a preference for relatively narrow networks during search. In general, this analysis would indicate that during evolution there was a particular focus on depth, with a tendency to find deeper networks which are relatively less topologically complex. Despite this, we found that some experiments were preferencing concatenation layers for certain runs. Future work could look at promoting concatenation layers and other branching operations through mutation and crossover.

\section{Conclusion}

Multi-branch topologies for the neuroevolution of DNNs can be a challenging prospect for distance-based surrogate models, when considering the structure of the architecture alone. In this work, we proposed the NeuroLGP multi-branch (NeuroLGP-MB) approach, that encodes a multi-branch topology and makes full use of semantic-based surrogate model by relying solely on the phenotypic information of the networks. As such, this method by-passed the need to encode complex topologies as found in genotypic based surrogate models. This new approach, has further been adapted with an update to the surrogate-assisted approach such that it used a pre-selection method to better initialise the surrogate model. We found that the new pre-selection variant produced the best networks on average for BreakHis $\times$40 and BreakHis $\times$200. Interestingly, Chest X-Ray dataset produced results that were similar across all models.
We found that in general, the new surrogate-PS was also robust like the original surrogate approach, while also maintaining a time advantage over the expensive model with an 11.8 - 16.6\% time reduction compared to the expensive model. An analysis of the depth and topological complexity revealed the preference for deeper networks that are not overly topologically complex. 

\begin{acks}
This publication has emanated from research conducted with the financial support of Taighde Éireann – Research Ireland under Grant number 18/CRT/6049. For the purpose of Open Access, the author has applied a CC BY public copyright licence to any Author Accepted Manuscript version arising from this submission. The simulations were performed on the Luxembourg national supercomputer MeluXina.
The authors gratefully acknowledge the LuxProvide teams for their expert support.
\end{acks}

\bibliographystyle{ACM-Reference-Format}
\bibliography{ref}


\end{document}